\documentclass[runningheads]{llncs}
\usepackage{graphicx}
\usepackage{multirow}
\usepackage{makecell}
\usepackage{etoolbox}
\usepackage{pgfplots}
\usepackage{caption}
\usepackage{subcaption}
\usepackage{filecontents}

\begin{document}
\title{Enhanced back-translation for low resource neural machine translation using self-training\thanks{Supported by NITDEF PhD Scholarship Scheme 2018.}}
\titlerunning{Enhanced back-translation using self-training}
%
\author{Idris Abdulmumin\inst{1,2}\orcidID{0000-0002-3795-8381} \and
Bashir Shehu Galadanci\inst{2} \and
Abubakar Isa\inst{1}}
\authorrunning{Abdulmumin et al.}
%
\institute{Ahmadu Bello University, Samaru Campus, Zaria, Kaduna State, Nigeria
\email{\{iabdulmumin, abubakarisa\}@abu.edu.ng} \and
Bayero University, Kano, Kano State Nigeria \\
\email{bsgaladanci.se@buk.edu.ng}}
\maketitle              
\begin{abstract}
Improving neural machine translation (NMT) models using the back-translations of the monolingual target data (synthetic parallel data) is currently the state-of-the-art approach for training improved translation systems. The quality of the backward system -- which is trained on the available parallel data and used for the back-translation -- has been shown in many studies to affect the performance of the final NMT model. In low resource conditions, the available parallel data is usually not enough to train a backward model that can produce the qualitative synthetic data needed to train a standard translation model. This work proposes a self-training strategy where the output of the backward model is used to improve the model itself through the forward translation technique. The technique was shown to improve baseline low resource IWSLT’14 English-German and IWSLT’15 English-Vietnamese backward translation models by 11.06 and 1.5 BLEUs respectively. The synthetic data generated by the improved English-German backward model was used to train a forward model which out-performed another forward model trained using standard back-translation by 2.7 BLEU.

\end{abstract}
\section{Introduction}
The neural machine translation (NMT) \cite{Bahdanau2014,Luong2015,Vaswani2017} is currently the simplest and yet the state-of-the-art approach for training improved translation systems \cite{Edunov2018,Ott2018}. They out-perform other statistical machine translation approaches if there exists a large amount of parallel data between the languages \cite{Koehn2017,Zoph2016}. Given the “right” amount of qualitative parallel data only, the models can learn the probability of mapping sentences in the source language to their equivalents in another language -- the target language \cite{Yang2019}. This “right” amount of qualitative parallel data is usually very large and, therefore, expensive to compile because it requires manual translation. The absence of large amounts of high-quality parallel data in many languages has led to various proposals for leveraging the abundant monolingual data that exists in either or both of the languages. These approaches include the self-training \cite{Specia2018}, forward translation \cite{Zhang2016}, back-translation \cite{Edunov2018,Sennrich2016a,Imamura2018,Caswell2019}, dual learning \cite{He2016} and transfer learning \cite{Zoph2016,Nguyen2017,Dabre2019,Kocmi2019}.

The back-translation has been used in current state-of-the-art neural machine translation systems \cite{Edunov2018,Lample2019,Lioutas2020}, outperforming other approaches in high resource languages and improving performance in low resource conditions \cite{Edunov2018,Hoang2018,Niu2018}. The approach involves training a target-to-source (backward) model on the available parallel data and using that model to generate synthetic translations of a large number of monolingual sentences in the target language. The available authentic parallel data is then mixed with the generated synthetic parallel data without differentiating between the two \cite{Sennrich2016a} to train a final source-to-target (forward) model. The quality of the forward translation model depends on the NMT architecture used in building the models \cite{Sennrich2016a}, the quality of the backward model \cite{Hoang2018,Zhang2018,Burlot2019}, the suitability of the synthetic data generation method used \cite{Edunov2018,Graca2019} and the ratio of the authentic data to the synthetic data \cite{Fadaee2018,Poncelas2018}. In low resource NMT, the authentic parallel data available is not sufficient to train a backward model that will generate qualitative synthetic data. Thus, various methods have been proposed to improve the quality of the backward model despite the lack of sufficient parallel data.

Hoang et al. \cite{Hoang2018} and Zhang et al. \cite{Zhang2018} used an iterative approach to enable the forward model to generate synthetic data that will be used to improve the backward model. Imamura et al. \cite{Imamura2018} suggest generating multiple synthetic sources through sampling given a target sentence. Niu et al \cite{Niu2018} trained a bilingual model for both the backward and forward translations and they reported improvement in low resource translations. Graca et al. \cite{Graca2019} proposed that selecting the most suitable synthetic data generation method will help reduce the inadequacies of the backward model. Dabre et al. \cite{Dabre2019} and Kocmi and Bojar \cite{Kocmi2019} proposed the use of a high-resource parent language pair through transfer learning to improve the backward model.

This work proposes the use of self-training -- also referred within the document as self-learning and forward translation -- \cite{Ueffing2006,Zhang2016,Specia2018} approach to improve the backward model. The output of the backward model -- which is ideally used with the authentic data to train the forward model in back-translation -- is used to improve the backward model itself. The self-training approach used is similar to that in \cite{Ueffing2006,Zhang2016,Specia2018} where a synthetic target-side data is used to improve the performance of the translation model instead of the synthetic source-side data in back-translation. But instead of using the approach to enhance the final model, we aim to enhance the backward model which then generates improved synthetic data for enhancing the final model. We also simplify the approach by removing the need for synthetic data quality estimation \cite{Specia2018} or freezing of training parameters \cite{Zhang2016}.

The work is similar to the iterative back-translation of Hoang et al. \cite{Hoang2018} and Zhang et al. \cite{Zhang2018}. The iterative back-translation requires the use of the monolingual source and target data to improve the backward and forward models respectively. The backward model generates synthetic sources to improve the forward model while the forward does the same for the backward model. This process is repeated iteratively until the required quality of translations are obtained. Instead, this work relies only on the monolingual target data to improve both models. Whereas the approaches above perform iterative back-translation to improve both models, our work uses forward translation (self-learning) to improve the backward model and back-translation to improve the forward model.

It was shown by Specia and Shah \cite{Specia2018} and Zhang and Zong \cite{Zhang2016} that using the monolingual source -- or the synthetic target -- data will potentially reduce the performance of the decoder. To mitigate this, Ueffing \cite{Ueffing2006} and Specia and Shah \cite{Specia2018} used quality estimation \cite{Specia2013} to determine the best-translated sentences to be used to retraining, while Zhang and Zong \cite{Zhang2016} proposed freezing the parameters of the decoder when training the model on the synthetic data. In this work, we showed that the self-learning approach is capable of improving a translation model even without synthetic data cleaning or freezing any learned parameters. We hypothesize that the amount of parallel data used in retraining the model is sufficient to improve the quality of the model if the model can differentiate between and learn effectively from the synthetic and natural data.

\subsection{Summary of Contributions}
We make the following contributions in this paper:
\begin{itemize}
\item instead of requiring the source and target data for improving the backward and forward models respectively, as in previous works, we investigated utilizing only the target-side monolingual data to improve both the backward and forward models in back-translation. Whereas the monolingual target data is used as the source data to improve the backward model (forward translation), we use the same data as the target data in the forward model training (back-translation). The work investigates different approaches for using the all of the synthetic data to improve the models.
\item we showed that even without data cleaning and/or freezing learned parameters, self-training improves the backward model; and that a forward model trained using the synthetic data generated from the improved backward model performs better than a forward model trained using the standard back-translation.
\item we showed that when a model -- backward or forward -- can differentiate between the authentic and synthetic data, it is able to utilize the quality in the authentic data and also, efficiently benefits from the increase in quantity resulted from adding the synthetic data.
\item we showed that the technique improves baseline low resource IWSLT’14 English-German and IWSLT’15 English-Vietnamese backward NMT models by 11.06 and 1.5 BLEUs respectively; and the synthetic data generated by the improved English-German backward model was used to train a forward model whose performance bettered that of a forward model trained using the standard back-translation technique by 2.7 BLEU.
\end{itemize}

\section{Related Works}
This section presents prior work on back-translation, forward translation and self-training.

\subsection{Back-Translation}

The use of monolingual data of target and/or source language has been studied extensively to improve the performance of translation models, especially in low resource settings. Gulcehre et al \cite{Gulcehre2017} explored the infusion of language models trained on monolingual data into the translation models. Currey et al. \cite{Currey2017} and Burlot and Yvon \cite{Burlot2019} proposed augmenting a copy or slightly modified copy of the target data as source respectively. Sennrich et al. \cite{Sennrich2016} and Zhang and Zong \cite{Zhang2016} proposed the back-translation and forward translation approaches respectively and He et al. \cite{He2016} used both source and target-side monolingual data to improve the translation models.

The back-translation approach has been shown to outperform other approaches in low and high resource languages \cite{Edunov2018,Hoang2018}. The quality of the models trained using back-translation depends on the quality of the backward model \cite{Edunov2018,Fadaee2018,Hoang2018,Burlot2019,Graca2019,Kocmi2019,Yang2019}. In low resource NMT, the authentic parallel data available is not sufficient to train a backward model that will generate qualitative synthetic data. To improve the quality of the synthetic parallel data, Hoang et al. \cite{Hoang2018}, Zhang et al. \cite{Zhang2018} and Caswell et al. \cite{Caswell2019} proposed the iterative back-translation -- iteratively using the back-translations of the source and target data to improve the backward and forward models respectively. Kocmi and Bojar \cite{Kocmi2019} and Dabre et al. \cite{Dabre2019} pre-trained a model using high resource languages and initialize the training of the low resource languages with the learned pre-trained weights – transfer learning. Niu et al. \cite{Niu2018} trained a bilingual system based on Johnson et al. \cite{Johnson2017} to do both forward and backward translations, eliminating the need for separate backward model. They reported improvement in low resource NMT.

\subsection{Forward Translation and Self-Training}

Forward translation (reverse back-translation, self-training or self-learning) was used to improve NMT \cite{Zhang2016} and other forms of statistical machine translation systems \cite{Ueffing2006,Specia2018}. Instead of the target-side monolingual data, forward translation uses the source-side monolingual data to improve the performance of a translation model. The available authentic data is used to train a source-to-target model. This model is then used to generate synthetic translations of the available (usually huge) source-side monolingual sentences. This data (synthetic target) is paired with the source-side data to create the synthetic parallel dataset. The resulting huge data is used to train a better source-to-target translation model. The synthetic data might contain mistakes that will likely reduce the performance of the models. Various works that used the forward translation (self-learning) approach proposed the use of other techniques to mitigate the effects of the noise present in the data, e.g. using quality estimation to automatically remove the sentences that are considered to be badly translated. Specia and Shah \cite{Specia2018} utilized an iterative approach to select the top n translations to retrain the generating model. Automatic quality estimation was used to determine sentences that are considered to be translated better than the others.

Ueffing \cite{Ueffing2006} explained self-training as an approach that takes the output of the machine translation model to improve the model itself. The work proposed the translation of monolingual source data, estimating the quality of the translated sentences, discarding those sentences whose quality is below a set threshold and subsequently training a new improved model on the mixed authentic and synthetic bilingual data. Zhang and Zong \cite{Zhang2016} proposed the forward translation (self-learning) to improve the encoder side of the NMT model. The authors suggested that back-translation improved the decoder by training it authentic target data and that when the NMT model is trained on authentic source data, the encoder will be improved. The use of synthetic data in back-translation may reduce the performance of the encoder because it is trained on the synthetic data. When using the synthetic target data in their approach, the authors tried to mitigate this problem by freezing the parameters of the decoder for the synthetic data during training.

\section{Methodology}

\subsection{Neural Machine Translation (NMT)}

This work is based on a unidirectional LSTM encoder-decoder architecture with Luong attention \cite{Luong2015}. This is a recurrent neural network RNMT architecture and it is summarized below. Our approach can be applied to other architectures such as the convolutional neural network NMT (CNMT) \cite{Gehring2017,Wu2019} and the Transformer \cite{Vaswani2017,Dehghani2019}.

Neural Machine Translation (NMT) is based on a sequence-to-sequence encoder-decoder system made of neural networks that models the conditional probability of a source sentence to a target sentence \cite{Bahdanau2014,Sutskever2014,Luong2015}. The encoder converts the input in the source language into a set of vectors while the decoder converts the set of vectors into the target language, word by word, through an attention mechanism -- introduced to keep track of context in longer sentences \cite{Bahdanau2014}. The NMT model produces the translated sentence by generating one target word at every time step. Given the “right” amount of qualitative parallel data only, the NMT model can learn the probability of mapping sentences in the source language to their equivalents in another language -- the target language -- word by word \cite{Yang2019}.

\begin{figure}
\centering
\includegraphics[clip, trim=4cm 18cm 0.5cm 5.5cm, width=1.50\textwidth]{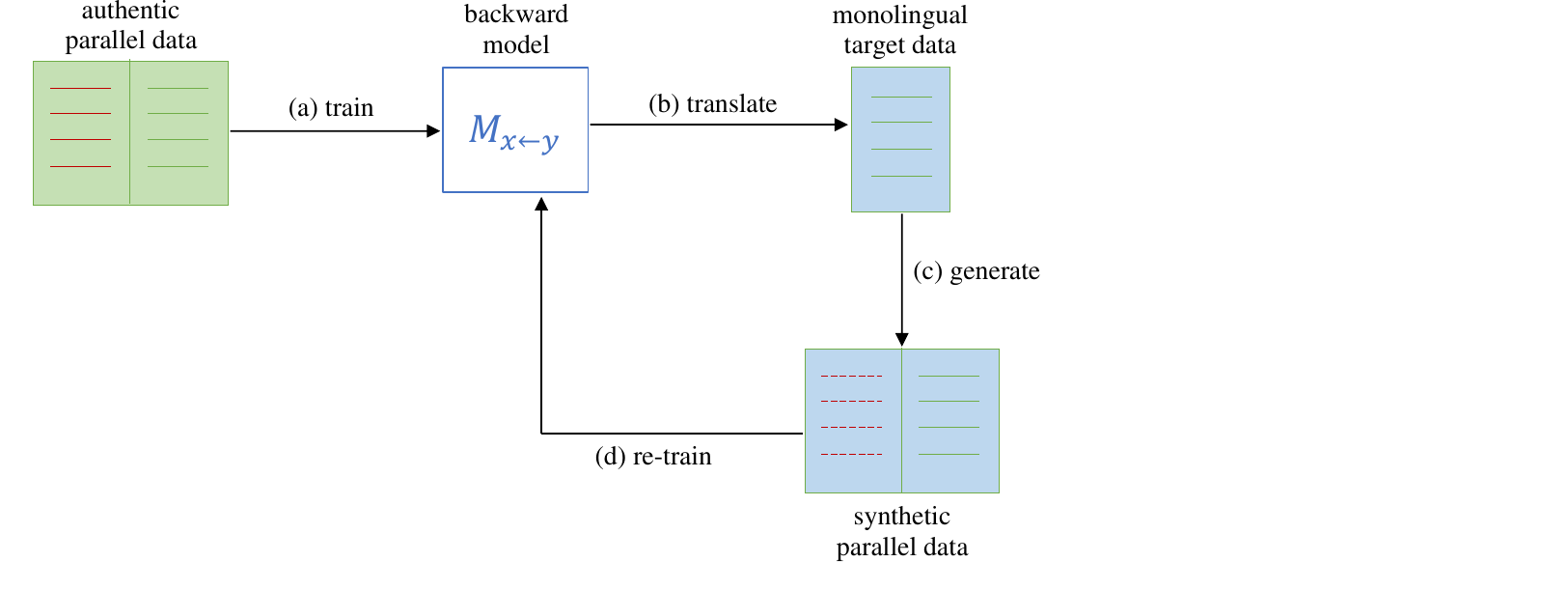}
\caption{Self-Learning for Improving the Backward Model: enabling the backward model to learn from the target language monolingual data}
\label{fig:sl}
\end{figure}

Given an input sequence \(X=(x_1,...,x_{T_x})\), the encoder -- made up of a bidirectional or unidirectional neural network with Long Short-Term Memory (LSTM) \cite{Hochreiter1997} or gated recurrent units (GRU) \cite{Cho2014} -- computes the annotation vector \(h_j\), which is a concatenation of the forward and backward hidden states \(\overrightarrow{h_j}\) and \(\overleftarrow{h_j}\) respectively. The decoder is made up of a recurrent neural network that takes a recurrent hidden state \(s_i\), the previously translated words \((y_1,...,y_{i-1})\) and a context vector \(c_i\) to predict the probability of the next word \(y_i\) as the weighted summation of the annotations \(h_j\). An alignment model -- a single layer feed-forward network which is learned jointly with the rest of the network through back-propagation -- which models the probability that \(y_i\) is aligned to \(x_i\) is used to compute the weight of each annotation \(h_j\).

All of the parameters in the NMT model, $\theta$, are optimized to maximize the following conditional log-likelihood of the M sentence aligned bilingual samples \[L(\theta) = \frac{1}{M}\sum_{m=1}^{M} \sum_{i=1}^{T_y}\log(p(y_i^m|y_{<i}^m, X^m,\theta))\].

\subsection{Overview of the Method}

As shown in \textbf{Algorithm 1}, given a set of parallel data and monolingual target sentences: \(D^P =\{(x^{(u)}, y^{(u)})\}_{u=1}^U\) and \(Y =\{(y^{(v)})\}_{v=1}^V\) respectively, we used the authentic parallel data: \(D^P\) to train a target-to-source model, \(M_{x \leftarrow y}\). This model -- the backward model -- is then used to translate the monolingual target data, \(Y\), to generate the synthetic parallel data: \(D^\prime =\{(x^{(v)}, y^{(v)})\}_{v=1}^V\). The resulting synthetic data is then used to improve the model either through fine-tuning it on the synthetic data, standard forward translation, tagged forward translation (similar to the tagged back-translation \cite{Caswell2019}) or through pre-training and fine-tuning \cite{abdulmumin2019tagless}. This technique is illustrated in Fig.~\ref{fig:sl}.

\begin{table}
\label{tab:algorithm}
\renewcommand{\arraystretch}{1.3}
\begin{tabular}{l}
\hline\noalign{\smallskip}
\textbf{ALGORITHM 1:} SELF-TRAINING \\
\noalign{\smallskip}\hline\noalign{\smallskip}
\makecell[tl]{\textbf{Input:} Parallel data \(D^P =\{(x^{(u)}, y^{(u)})\}_{u=1}^U\) and Monolingual \\ \quad \quad \quad \quad target data \(Y =\{(y^{(v)})\}_{v=1}^V\)} \\
\noalign{\smallskip}
1: \textbf{procedure} SELF-TRAINING \\
2: \quad Train backward model \(M_{x \leftarrow y}\) on bilingual data \(D^P\) \\
3: \quad Let \(D^\prime = \) synthetic parallel corpora generated for \(Y\) using \(M_{x \leftarrow y}\); \\
4: \quad Train improved backward model \(M_{x \leftarrow y}^+\) on bilingual data \(D^P \cup D^\prime\); \\
5: \textbf{end procedure} \\
6: \textbf{procedure} BACK-TRANSLATION \\
7: \quad Let \(D^\ast = \) synthetic parallel corpora generated for \(Y\) using \(M_{x \leftarrow y}^+\); \\
8: \quad Train forward model \(M_{x \rightarrow y}\) on bilingual data \(D^P \cup D^\ast\); \\
9: \textbf{end procedure} \\
\noalign{\smallskip}
\textbf{Output:}  improved \(M_{x \leftarrow y}\) and \(M_{x \rightarrow y}\) models\\
\noalign{\smallskip}\hline
\end{tabular}
\end{table}

Previous works that used self-training to improve machine translation models (e.g. \cite{Specia2018}, \cite{Ueffing2006}) proposed an extra step of data cleaning or freezing parameters (not updating the parameters of the decoder when training on the synthetic target data) to achieve the required performance. Our approach does not require any specialized approach of data cleaning or training regime. We showed that the simple act of joining the synthetic and authentic data can improve the model. We went further to show that when the backward model can differentiate between the synthetic data and authentic data, the performance increases even further. We investigated pre-training and fine-tuning, and tagging as methods that will help the model differentiate between the data. Also, we used self-training in this work only to enhance the backward model in the back-translation approach rather than training a final translation model.

\subsection{Data}

In this work, we used the data from the IWSLT 2014 German-English shared translation task \cite{Cettolo2014}. For pre-processing, we used the data cleanup and train, dev and test split in Ranzato et al. \cite{Ranzato2016}, resulting in 153,348, 6,970 and 6,750 parallel sentences for training, development and testing respectively. For the second low resource dataset, we used the pre-processed low resource English-Vietnamese parallel data \cite{Luong2015} of the IWSLT 2015 Translation task \cite{Cettolo2012}. We then utilized the 2012 and 2013 test sets for development and testing respectively. Table ~\ref{tab1} shows the data statistics. We used 400,000 English monolingual sentences of the pre-processed \cite{Luong2015} WMT 2014 English-German translation task \cite{Bojar2017} for the monolingual data. We learned byte pair encoding (BPE) \cite{Sennrich2016b} with 10,000 merge operations on the training dataset, applied it on the train, development and test datasets and, afterwards, build the vocabulary on the training dataset.

\begin{table}[!ht]
\renewcommand{\arraystretch}{1.3}
\caption{Data Statistics}
\label{tab1}
\begin{tabular}{|c|c|c|c|c|c|}
\hline
\multirow{2}{*}{data} & \multicolumn{3}{c|}{train} & \multirow{2}{*}{dev} & \multirow{2}{*}{test} \\ \cline{2-4}
 & sentences & \multicolumn{2}{c|}{words (vocab)} &  &  \\ \hline
\multirow{2}{*}{IWSLT’14 En-De} & \multirow{2}{*}{153, 348} & En & De & \multirow{2}{*}{6, 970} & \multirow{2}{*}{6, 750} \\ \cline{3-4}
 &  & 2,837,240 (50,045) & 2,688,387 (103,796) &  &  \\ \hline
\multirow{2}{*}{IWSLT’15 En-Vi} & \multirow{2}{*}{133, 317} & En & Vi & \multirow{2}{*}{1, 553} & \multirow{2}{*}{1, 268} \\ \cline{3-4}
 &  & 2,706,255 (54,169) & 3,311,508 (25,615) &  &  \\ \hline
\makecell[tl]{WMT’14 En-De –\\ Monolingual English} & 400, 000 & \multicolumn{2}{c|}{9,918,380 (266,640)} & - & - \\ \hline
\end{tabular}
\end{table}

\subsection{Set-up}

We used the NMTSmallV1 configuration of the OpenNMT-tf \cite{Klein2017}, the TensorFlow \cite{Abadi2016} implementation, a framework for training NMT models. The configuration is a 2-layer unidirectional LSTM encoder-decoder model with Luong attention \cite{Luong2015} with 512 hidden units and a vocabulary size of 50,000 for both source and target languages. The optimizer we used is Adam \cite{kingma2014adam}, a batch size of 64, a dropout probability of 0.3 and a static learning rate of 0.0002. The models are evaluated on the development set after every 5,000 training steps. For evaluation, we used the bi-lingual evaluation understudy (BLEU) \cite{Papineni2002}.

Training is stopped when the models reach a total of 200,000 training steps or when there is no improvement of over 0.2 BLEU after the evaluation of four consecutive training steps. We used this set-up to train all the models and unless stated otherwise: (1) there was no extra training for any model after either of the stopping criteria were met; (2) we average the last 8 checkpoints of every model trained to obtain a better performance and; (3) we update the vocabulary of every checkpoint with the that of the new training data before fine-tuning.

\section{Experiments and Results}
\label{result}

First, we train a backward model (En-De) -- \emph{baseline} -- for 80,000 training steps, achieving the best score of 10.03 BLEU after 65,000 training steps. Averaging the last 8 checkpoints results in a better performance of 10.25 BLEU and we used this average checkpoint as our  backward model for generating the synthetic data. The resulting parallel data is labelled as \emph{synth-A}. We then used the authentic parallel data and \emph{synth-A} to train an improved backward model. Apart from the standard forward translation (self-learning) technique of mixing the data and training from the scratch, we followed other training strategies to enable the model to differentiate between the authentic and synthetic parallel data. The results obtained by using these various strategies are shown in Table~\ref{tab2}.

\subsection{Forward translation}

\begin{table}[!ht]
\renewcommand{\arraystretch}{1.3}
\centering
\caption{Scores for best checkpoints and checkpoint averaging of backward models trained using different techniques.}
\label{tab2}
\begin{tabular}{|c|c|c|c|c|c|c|}
\hline
\multirow{2}{*}{} & \multirow{2}{*}{} & \multicolumn{3}{c|}{self-training (this work)} \\ \cline{3-5} 
 & \begin{tabular}[c]{@{}c@{}}baseline \\ \cite{Luong2015,Sennrich2016} \end{tabular} & backward\_ft & tagged\_ft & \begin{tabular}[c]{@{}c@{}}pre-train \& \\ fine-tune\end{tabular} \\ \hline
\begin{tabular}[c]{@{}c@{}}best score \\ (training step)\end{tabular} & \begin{tabular}[c]{@{}c@{}}10.03 \\ (65k)\end{tabular} & \begin{tabular}[c]{@{}c@{}}20.48 \\ (150k)\end{tabular} & \begin{tabular}[c]{@{}c@{}}20.72\\    (150k)\end{tabular} & \begin{tabular}[c]{@{}c@{}}20.77 \\ (115k)\end{tabular} \\ \hline
average & 10.25 & 20.98 & 21.02 & 20.35 \\ \hline
\end{tabular}
\end{table}

We mixed the authentic parallel data and \emph{synth-A} without differentiating between the two and trained the backward model from scratch. The model trained for 180,000 steps before stopping. The best score obtained was more than double the performance of \emph{baseline} with an improvement of 10.45 BLEU. The averaged checkpoint -- \emph{backward\_ft} -- gained an improvement of 10.73 BLEU over \emph{baseline}. This huge improvement supports the hypothesis that even without data cleaning and/or freezing of decoder parameters, the model is able to learn from the synthetic data generated by itself. After a few training steps, the performance of \emph{backward\_ft} started to improve significantly over \emph{baseline} (see Fig.~\ref{fig2}).

\subsection{Tagged forward translation}

To enable the backward model to differentiate between the two data, we experimented the ‘tagged forward translation’ -- coined from the tagged back-translation of \cite{Caswell2019}. While they used the ‘\emph{\textless BT\textgreater}’ tag to indicate if a source was synthetic, we instead utilized the ‘\emph{\textless SYN\textgreater}’ tag to differentiate between authentic and synthetic target sentences. We named the model that was trained using this approach as \emph{tagged\_ft}. Although the tagged approach outperforms the standard approach, the difference observed in the performances of the \emph{tagged\_ft} and \emph{backward\_ft} models was not significant.

\subsection{Pre-training and Fine-tuning}

\begin{table}[!ht]
\renewcommand{\arraystretch}{1.3}
\centering
\caption{Improvements observed after re-training the backward model on the synthetic target data for English Vietnamese machine translation.}
\label{tab3}
\begin{tabular}{|c|c|c|c|}
\hline
\multirow{2}{*}{} &  & \multicolumn{2}{c|}{self-training} \\ \cline{3-4} 
 & baseline & \begin{tabular}[c]{@{}c@{}}forward translation\end{tabular} & \begin{tabular}[c]{@{}c@{}}pre-training \& fine-tuning\end{tabular} \\ \hline
\begin{tabular}[c]{@{}c@{}}best score (training step)\end{tabular} & \begin{tabular}[c]{@{}c@{}}24.78 (50k)\end{tabular} & \begin{tabular}[c]{@{}c@{}}25.97 (105k)\end{tabular} & \begin{tabular}[c]{@{}c@{}}26.22 (125k)\end{tabular} \\ \hline
average & 25.79 & 26.38 & 27.29 \\ \hline
\end{tabular}
\end{table}

Following the work of Abdulmumin et al. \cite{abdulmumin2019tagless}, we trained the models using the following approache: pre-training on the synthetic data and fine-tuning on the authentic data. we We experimented mixing the authentic data and \emph{synth-A} to learn joint BPE and build a vocabulary of the mixed data. Afterwards, we pre-trained the backward model on \emph{synth-A} and fine-tuned it on the authentic data. The performance of the average of the last 8 checkpoints was a little bit lower (-0.87 BLEU) than that of the other strategies, but the best checkpoints in the strategy outperformed the others. We realized that averaging the last 8 checkpoints hurts the performance because continuing to train the model after 145,000 training steps produced poor checkpoints (see Fig.~\ref{fig2}). We, instead, took the average of the previous 8 checkpoints starting from the checkpoint at 145,000 training steps. This resulted in an increased performance of the model to 21.31 BLEU (+0.96), an increase of 0.1 BLEU over the previous approach. This appears to have the best performance among the models trained so far. We, therefore, used this model to generate \emph{synth-B} -- a synthetic parallel data generated for the monolingual sentences.

\subsection{English Vietnamese (En-Vi)}

We used the En-Vi dataset to test the results obtained on the En-De dataset. A backward model was trained using the English-Vietnamese parallel data for 55,000 training steps. The model (En-Vi) achieved a BLEU score of 24.78 after 50,000 training steps. An average of the last 8 checkpoints resulted in an improved performance of 25.79 BLEU and the checkpoint was labelled \emph{envi\_baseline}. The model, \emph{envi\_baseline}, was used to translate the monolingual English data to generate the synthetic parallel data -- \emph{synth-C}. The authentic data was mixed with \emph{synth-C} to train a backward model -- \emph{envi\_backward} -- from the scratch. The model gained a +1.19 BLEU (see Table~\ref{tab3}) on the best checkpoint and 0.59 BLEU on the average checkpoint over \emph{envi\_baseline}. The results are shown in Table~\ref{tab3}. We then used the pre-training and fine-tuning approach to train the backward model. Even during the pre-train stage of this approach, the average checkpoint achieved a performance that is close to that of \emph{envi\_baseline} -- a score of 24.82 (-0.97) BLEU. This supports the claim by \cite{Edunov2018} that training a translation model on the synthetic parallel data only can reach a performance similar to the model that is trained on authentic data only.

\begin{table}[!ht]
\renewcommand{\arraystretch}{1.3}
\caption{Forward models (De-En) trained using different quality of synthetic data.}
\label{tab4}
\begin{tabular}{|c|c|c|c|c|c|}
\hline
\multirow{2}{*}{} & \multirow{2}{*}{\begin{tabular}[c]{@{}c@{}}baseline\\ forward\\ model \cite{Luong2015} \end{tabular}} & \multicolumn{2}{c|}{\begin{tabular}[c]{@{}c@{}}baseline backward model \cite{Sennrich2016}\\ (10.25 BLEU)\end{tabular}} & \multicolumn{2}{c|}{\begin{tabular}[c]{@{}c@{}}self-trained backward model\\   (this work; 21.22 BLEU)\end{tabular}} \\ \cline{3-6} 
 &  & \begin{tabular}[c]{@{}c@{}}standard back-\\ translation\end{tabular} & \begin{tabular}[c]{@{}c@{}}pre-training \&\\ fine-tuning\end{tabular} & \begin{tabular}[c]{@{}c@{}}standard back-\\ translation\end{tabular} & \begin{tabular}[c]{@{}c@{}}pre-training \&\\ fine-tuning\end{tabular} \\ \hline
\begin{tabular}[c]{@{}c@{}}best score\\ (training step)\end{tabular} & \begin{tabular}[c]{@{}c@{}}20.30\\ (75k)\end{tabular} & \begin{tabular}[c]{@{}c@{}}25.11\\ (150k)\end{tabular} & \begin{tabular}[c]{@{}c@{}}25.32\\ (115k)\end{tabular} & \begin{tabular}[c]{@{}c@{}}27.41\\ (110k)\end{tabular} & \begin{tabular}[c]{@{}c@{}}28.38\\ (135k)\end{tabular} \\ \hline
average & 20.95 & 25.87 & 26.03 & 27.87 & 28.73 \\ \hline
\end{tabular}
\end{table}

We observed that although the quality of the synthetic data determines the feasibility of the claim, it is true for either synthetic target or source data. The performance of the backward model that was pre-trained on the synthetic data generated by \emph{baseline} (Section~\ref{result}) -- which was in itself poor (10.25 BLEU) -- was significantly less than the that of the baseline (-2.83 BLEU). After fine-tuning, the performance of the model improved to 27.29 (+1.5). Although some gain in performance was realized, the difference was not as significant as it was observed on the En-De dataset -- +1.5 on En-Vi compared to +9.1 on En-De. This may have been because the backward model, \emph{envi\_baseline}, was already good compared to \emph{baseline}.

\subsection{Back-Translation}

It is expected, as shown in many studies (e.g. \cite{Edunov2018}, \cite{Poncelas2018}), that a better synthetic data generated using a good backward model will result in an improved forward model. We used the outputs of the backward models -- \emph{synth-A} and \emph{synth-B} -- to train final forward models. We expected the quality of \emph{synth-B} to be better since it was generated using the best backward model among those trained in the experiments above. Both of the models trained using the standard back-translation and the pre-training and fine-tuning approaches performed better than the models trained using the same approaches but with \emph{synth-A} (see Fig. ~\ref{fig3}).

Table ~\ref{tab4} shows the performance of the models trained: without synthetic data; with \emph{synth-A} and; with \emph{synth-B}. The best model was obtained through pre-training and fine-tuning using authentic data and \emph{synth-B}. The model out-performed the baseline forward model by a BLEU score of 7.78 (28.73 BLEU). Although using \emph{synth-A} improved the performance of the forward model over the baseline (+4.92 and +5.08 BLEUs using standard back-translation and pre-training and fine-tuning respectively), the effect of the backward model self-training ensured that the quality of \emph{synth-B} was superior and the model trained using this data improved the forward model further by over +2 BLEU.

\begin{figure}
\centering
\begin{minipage}[t]{0.48\textwidth}
\resizebox{0.95\linewidth}{!}{%
\begin{tikzpicture}
\begin{axis}[
    xlabel={Training steps (thousands)},
    ylabel={BLEU},
    xmin=0, xmax=180,
    ymin=0, ymax=25,
    xtick={0,30,60,90,120,150,180},
    ytick={0,5,10,15,20,25},
    legend pos=south east,
    legend cell align={left},
    ymajorgrids=true,
    grid style=dashed,
]

\addplot[
    color=blue,
    mark=square,
    ]
    table {baseline.data};
    \addlegendentry{\emph{baseline}}

\addplot[
    color=red,
    mark=o,
    ]
    table {ft.data};
    \addlegendentry{\emph{forward\_ft}}
    
\addplot[
    color=black,
    mark=triangle,
    ]
    table {tagged.data};
    \addlegendentry{\emph{tagged\_ft}}

\addplot[
    color=green,
    mark=star,
    ]
    table {pretrainC.data};
    \addlegendentry{\emph{P\&F}}
    
\end{axis}
\end{tikzpicture}
}
\captionof{figure}{Performance of \emph{baseline} backward model compared to self-trained backward models using tagging, standard forward translation and the pre-train and fine-tune approaches. NOTE: P\&F means Pre-train and Fine-tune}
\label{fig2}
\end{minipage}%
\hfill
\begin{minipage}[t]{0.48\textwidth}
\resizebox{0.95\linewidth}{!}{%
\raggedleft
\begin{tikzpicture}
\begin{axis}[
    xlabel={Training steps (thousands)},
    ylabel={BLEU},
    xmin=0, xmax=190,
    ymin=0, ymax=30,
    xtick={0,30,60,90,120,150,180},
    ytick={0,5,10,15,20,25,30},
    legend pos=south east,
    legend cell align={left},
    ymajorgrids=true,
    grid style=dashed,
]

\addplot[
    color=blue,
    mark=square,
    ]
    table {baseline.data};
    \addlegendentry{\emph{A}}

\addplot[
    color=red,
    mark=star,
    ]
    table {forwardbt.data};
    \addlegendentry{\emph{B}}
    
\addplot[
    color=red,
    mark=triangle,
    ]
    table {forwardpretrain.data};
    \addlegendentry{\emph{C}}
    
\addplot[
    color=black,
    mark=star,
    ]
    table {forwardimpbt.data};
    \addlegendentry{\emph{D}}
    
\addplot[
    color=black,
    mark=triangle,
    ]
    table {forwardimppretrain.data};
    \addlegendentry{\emph{E}}
    
\end{axis}
\end{tikzpicture}
}
\captionof{figure}{Forward models (De-En) trained using different quality of synthetic data. KEY: A = \emph{baseline}, B = \emph{back-translation}, C = \emph{pre-train \& fine-tune}, D = \emph{improved back-translation}, E = \emph{improved pre-train \& fine-tune}}
\label{fig3}
\end{minipage}
\end{figure}

\section{Discussion}

Neural machine translation systems suffer when trained on scanty data - low resource languages. Back-translation is an approach that was introduced to improve the performance of these and other category of languages. But various studies have shown that in low resource set-ups, the performance require other special approaches to reach an acceptable standard for translation quality. This work, therefore, proposes a new method of using the target-side monolingual data more effectively to improve the performance of the back-translation approach. Whereas the back-translation was used to specifically improve the forward model, we used the self-training approach through forward translation to improve, also, the performance of the backward model. The method performed very well on low resource English-German and English-Vietnamese languages and can be applied to any other low resource neural machine translation. The method can be investigated also in high resource languages.

We investigated various approaches such as the forward translation, tagged forward translation and various pre-training and fine-tuning strategies with the later two implemented to enable the model differentiate between synthetic and authentic parallel data during training. We observed that the proposed method out-performed the backward model in standard back-translation. It was claimed in \cite{Specia2018} and \cite{Zhang2016} that the model's performance may be affected when using self-training because of the noise in the synthetic data. Instead, we found that providing a means for the model to differentiate between synthetic and authentic parallel data is just sufficient for the self-training method to perform as desired. Even though the self-training is by itself successful at improving the model, using tags or pre-training and fine-tuning have shown to improve the model's performance.

\begin{table}[!ht]
\renewcommand{\arraystretch}{1.3}
\caption{A German-to-English translation example in the IWSLT-DE 14 test set.}
\label{tab5}
\begin{tabular}{p{3cm}|p{9cm}}
\hline \hline
Source & und es funktionierte . wieder hatten wir etwas magisches geschaffen . und die wirkung im publikum war dieselbe . allerdings haben wir mit dem film schon ein bisschen mehr geld eingespielt \\ \hline

Reference & and it did , and we created magic again , and we had the same result with an audience -- although we did make a little more money on that one. \\ \hline

Baseline & and it worked . again . we had something magical , and the effect in the audience was the same thing , but we had a little more compound with a little more money. \\ \hline

Standard BT & and it worked . again , we created something magical , and the effect in the audience was the same , but we had a little bit more money on the film. \\ \hline

Pre-train \& Fine-tune & and it worked . again , we created something magical , and the effect in the audience was the same thing , but we had a little bit more money in the movie. \\ \hline

Improved Standard BT & and it worked . again , we created something magical , and the effect in the audience was the same . but we have a little bit more money. \\ \hline

Improved Pre-train \& Fine-tune & and it worked . again , we had created something magical , and the effect in the audience was the same . but we did a little bit more with the movie. \\ \hline

\end{tabular}
\end{table}

The work was evaluated on the low resource IWSLT 14 English-German translation. We also used the IWSLT 15 English-Vietnamese parallel data to confirm the positive results obtained using the approach. In Table 5, we showed a sample translation from English to German. Our improved model was able to produce exact translation to most  of the referenced translation: ''... wir 3 milliarden stunden pro woche mit online-spielen'' and the other part where the translation generated was different, the meaning was the same: ''derzeit verbringen'' 'vs' ''im moment geben''. The self-trained model was also able to generate exact translation to most of the referenced text but the model could only specify the adverb ''now'' instead of the referenced ''right now''. The improved model (trained using the best pre-train and fine-tune approach) generated ''at the moment'' which was a better equivalent to the next best translation system. Though the rest of the models could not perform better than the two discussed, the quality superiority of our approach can be seen on the models trained. For the forward model, the effects of the improved models were observed in their performances. In Table 6, we also translated a given German source text to English. The performances of the last two models (trained on the synthetic sentences generated by the backward model improved using our approach), and especially the last model, were superior than the rest of the other models.

The pre-training and fine-tuning approach has shown to be the better approach when applying the method we proposed in this work. Unlike in \cite{abdulmumin2019tagless}, we investigated different approaches that will suit better for our approach. We found that pre-training first on the synthetic data and thereafter fine-tuning the model on the authentic data is the best strategy. Fine-tuning on the synthetic data was found to hurt the model. This can be attributed to the lack of quality in the synthetic data used for fine-tuning compared to the authentic data used during pre-training, supporting the same claim in the work of \cite{abdulmumin2019tagless} that fine-tuning on the synthetic data does not improve performance, it only hurts it.

\section{Conclusion \& Future work}

To the best of our knowledge, this is the first work that investigated an all-round utilization of the synthetic data to improve neural machine translation especially on low resource languages. These category of languages lag their high resource counterparts even if the same methods for improving their performance are applied. The back-translation has been shown to improve translation performance across board but in low resource languages, the performance is still less than desirable. We applied joint backward and forward translation to utilize the target-side monolingual data in improving the performance of neural machine translation systems in low resource languages. Experimental results obtained on English-German and English-Vietnamese have shown that the approach is superior to that of the widely successful back-translation approach. The approach is straightforward and can be applied on any low resource language translation to achieve a better and more acceptable translation performance. It could also be applied on high resource languages to improve the performance.

We showed that the approach is capable of improving the performance of the model even without using specialized data cleaning methods such as quality estimation. We also showed that the quality of the backward model is improved when the model can differentiate between the two data. This is also true for all models trained on synthetic and authentic data as shown in the training of the forward models. The work can be extended by comparing the performance of the proposed method with the other implementations of the self-learning approach when improving the backward model. Repeated retraining of the backward model -- iterative self-training -- can be explored in future works to determine the extent to which the backward model’s output can be used to improve itself. We also intend to investigate the efficacy of the approach on high resource languages.

%
%
%
%

\end{document}